\documentclass[sigconf, nonacm]{acmart}
\usepackage{multirow}
\usepackage[table]{xcolor}
\newcommand{\gainColor}[2]{\cellcolor{orange!#1}#2}
\AtBeginDocument{%
  }

\begin{document}

\title{Robust Mesh Saliency Ground Truth Acquisition in VR via View Cone Sampling and Manifold Diffusion}

\author{Guoquan Zheng}
\authornote{Both authors contributed equally to this research.}
\author{Jie Hao}
\authornotemark[1]
\affiliation{%
  \institution{Shanghai Jiao Tong University}
  \city{Shanghai}
  \country{China}}

\author{Huiyu Duan}
\affiliation{%
  \institution{Shanghai Jiao Tong University}
  \city{Shanghai}
  \country{China}}

\author{Long Tang}
\affiliation{%
  \institution{Shanghai Jiao Tong University}
  \city{Shanghai}
  \country{China}}

\author{Shuo Yang}
\affiliation{%
  \institution{Beijing University of Chemical Technology}
  \city{Beijing}
  \country{China}}

\author{Yucheng Zhu}
\affiliation{%
  \institution{Shanghai Jiao Tong University}
  \city{Shanghai}
  \country{China}}

\author{Yongming Han}
\affiliation{%
  \institution{Beijing University of Chemical Technology}
  \city{Beijing}
  \country{China}}

\author{Liang Yuan}
\affiliation{%
  \institution{Shanghai Jiao Tong University}
  \city{Shanghai}
  \country{China}}

\author{Patrick Le Callet}
\affiliation{%
  \institution{Nantes University}
  \city{Nantes}
  \country{France}}

\author{Guangtao Zhai}
\affiliation{%
  \institution{Shanghai Jiao Tong University}
  \city{Shanghai}
  \country{China}}








\begin{abstract}
As the complexity of 3D digital content grows exponentially, understanding human visual attention is critical for optimizing rendering and processing resources. Therefore, reliable 3D mesh saliency ground truth (GT) is essential for human-centric visual modeling in virtual reality (VR). 
However, existing VR eye-tracking frameworks are fundamentally bottlenecked by their underlying acquisition and generation mechanisms. The reliance on zero-area single ray sampling (SRS) fails to capture contextual features, leading to severe texture aliasing and discontinuous saliency signals. And the conventional application of Euclidean smoothing propagates saliency across disconnected physical gaps, resulting in semantic confusion on complex 3D manifolds.
This paper proposes a robust framework to address these limitations. We first introduce a view cone sampling (VCS) strategy, which simulates the human foveal receptive field via Gaussian-distributed ray bundles to improve sampling robustness for complex topologies. 
Furthermore, a hybrid Manifold-Euclidean constrained diffusion (HCD) algorithm is developed, fusing manifold geodesic constraints with Euclidean scales to ensure topologically-consistent saliency propagation. 
We demonstrate the improvement in performance over baseline methods and the benefits for downstream tasks through subjective experiments and qualitative and quantitative methods. By mitigating “topological short-circuits” and aliasing, our framework provides a high-fidelity 3D attention acquisition paradigm that aligns with natural human perception, offering a more accurate and robust baseline for 3D mesh saliency research.
\end{abstract}

\begin{teaserfigure}
    \centering
    \includegraphics[width=\textwidth]{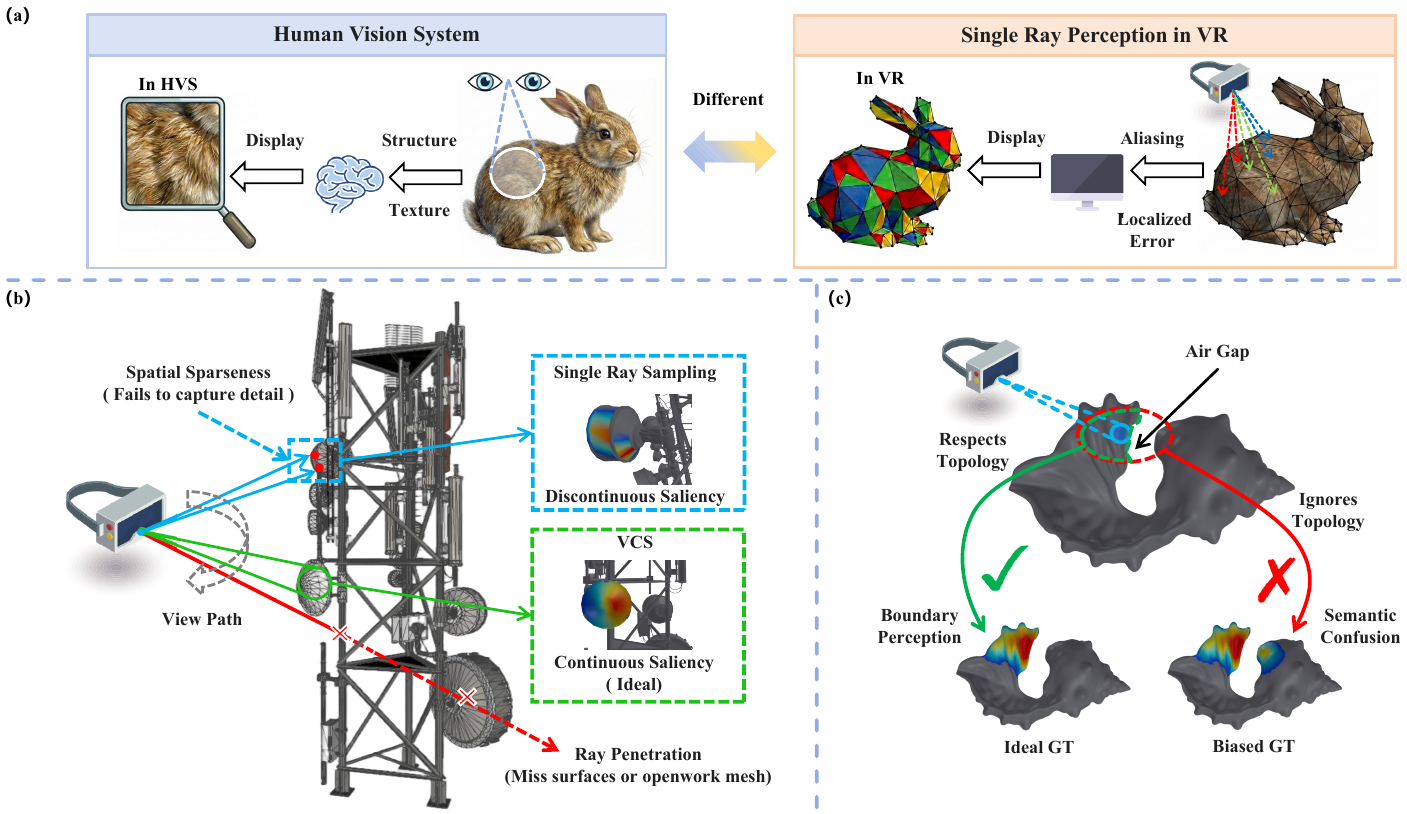}
    \caption{(a) Discrepancy between perceptual mechanism and SRS method. (b) Sparse geometric structure may introduce significant discontinuous saliency or penetrating accidentally ray collision. (c) Ignoring the obstacle of geometric gaps on visual attention.}
    \label{fig1}
\end{teaserfigure}

\maketitle

\section{Introduction}
\label{sec:intro}
With the rapid evolution of Virtual Reality (VR), Augmented Reality (AR) and the Metaverse, immersive multimedia applications are reshaping human perception of the digital world at an unprecedented pace\cite{bastug2017toward, duan2022saliency, liu2025moa, yang2025odi}. Serving as the fundamental representation for constructing virtual environments, the 3D colored mesh model has emerged as an indispensable data format for immersive experiences. This is largely attributed to its ability to simultaneously and accurately characterize both intricate geometric structures and rich texture appearances\cite{hao2025mm, hao2025mixed}.

To facilitate the processing of massive 3D data under constrained computational and transmission resources, mesh saliency prediction has emerged as a critical technology \cite{lee2005mesh}. By simulating the Human Visual System (HVS) attention mechanism\cite{itti1998model}, this technique identifies visually significant regions on 3D surfaces, providing a foundation for various downstream tasks such as mesh simplification \cite{zhou2016view}, view-dependent rendering \cite{weier2016foveated}, geometry compression, and perceptual quality assessment \cite{tang2024dtsn,tang2024fspn,zheng2023review, duan2022confusing, yang2025quality, li2025r, zhang2024embodied}.

However, developing robust human-centric saliency models requires high-quality ground truth (GT) data \cite{lavoue2018visual}. However, it is particularly challenging to acquire attention data for 3D meshes compared to traditional 2D images due to the complex geometric structure (such as hollow shape) and omnidirectional viewing method \cite{ding2023towards, zhu2021viewing, zhu2025does}. Therefore, establishing a rigorous subjective experimental paradigm to acquire fine grained saliency GT is primary for advancing both prediction models and perceptual applications\cite{song2021mesh, duan2025finevq, zhu2025future}.

Early research acquires the GT of mesh saliency by collecting manually marked points of interest on 3D objects to reflect the distribution of surface saliency density. 
However, in such approaches, subjects tend to make selections based on semantic understanding rather than visual saliency triggered by the visual stimuli themselves. 
Furthermore, the operation of manually marking vertices introduces additional human-computer interaction overhead, which is an unnatural discrete selection mode rather than the continuous process of visual exploration\cite{chen2012schelling,dutagaci2012evaluation}. 
Subsequent methods project 3D mesh models into 2D views as visual stimuli. 
These methods use screen-based eye trackers to capture fixation points, map the 2D coordinates back onto the 3D model, and apply Gaussian filters to smooth the fixations, thereby generating vertex-based saliency maps. 
However, this GT construction method lacks critical 3D depth cues, such as binocular disparity. 
Moreover, the saliency distribution derived from planar images and the actual visual attention distribution in the real 3D space has fundamental discrepancy\cite{lavoue2018visual}.

With recent advancements in VR technology, collecting eye-tracking data for 3D mesh models within VR environments has emerged as a mainstream solution. 
These methods typically acquire eye-tracking data by determining gaze positions through the collision of a ray emitted from the viewpoint with the model\cite{ding2023towards,martin2024sal3d,zhang2025mesh,zhang2025textured, zhang2025unified, zhang2025elevating}.
Then, the filtered fixation points are smoothed using cone-shaped beams with a Gaussian distribution to produce the final visual saliency map.
Compared to early approaches, VR environments accurately reproduce the spatial structure and depth information of 3D mesh models. Subjects can move freely within the VR space for visual exploration, thereby avoiding subjective biases introduced by additional manual operations. 
Furthermore, the implicit recording of eye-tracking data intuitively reflects the subjects' most instinctive interest distribution in a natural state. However, in-depth research has revealed that these VR-based mesh saliency acquisition frameworks still share several common limitations:

\textbf{(i) Discrepancy between perceptual mechanism and single ray sampling (SRS) method.} 
The HVS integrates texture and structural cues via receptive fields. However, as a zero-area discrete sampling method, SRS induces aliasing when encountering high-frequency textures. This mechanism leads to recording biases in the contextual perception of local salient patterns and geometric features as shown in Figure. \ref{fig1}(a) \cite{mcnamara2010perceptually}.

\textbf{(ii) Sparse geometric structure may introduce significant discontinuous saliency or penetrating accidentally ray collision.} 
Existing single ray methodologies predominantly focus on low-poly models with simple topologies. On high-resolution meshes, filtering on single ray within limited mesh surfaces significantly leads to discontinuous saliency. Furthermore, the accident ray penetration effect in non-manifold geometries triggers error attention. These factors compromise the accurate modeling of saliency density on complex mesh surfaces as shown in Figure. \ref{fig1}(b) \cite{zhang2025textured}.

\textbf{(iii) Ignoring the obstacle of geometric gaps on visual attention.} 
Conventional post-processing relies on Euclidean-based smoothing that disregards the intrinsic topological properties of the 3D mesh manifold. For geometries containing gaps, the Gaussian kernel propagates directly across disconnected spatial voids. This failure to respect physical boundaries weakens the topological independence of surface regions and introduces semantic confusion into the generated GT as shown in Figure. \ref{fig1}(c) \cite{jeong2017saliency}.

To address these challenges, we propose a robust VR-based framework for 3D mesh saliency GT construction, facilitating precise attention modeling on complex textures and topologies. We establish an immersive VR scenario to enable natural exploratory observation. In the acquisition phase, we propose a Gaussian-distributed \textbf{V}iewing \textbf{C}one \textbf{S}ampling (VCS) strategy to mitigate discreteness and aliasing inherent in SRS. By emitting a Gaussian ray bundle to simulate foveal receptive fields, VCS expands isolated fixations into weighted gaze regions, which significantly enhances robustness against complex textures and noise. For GT construction, we propose a \textbf{H}ybrid Manifold-Euclidean \textbf{C}onstraint \textbf{D}iffusion (HCD) algorithm that fuses manifold structures with Euclidean scales to overcome adjacency confusion caused by traditional smoothing. Our pipeline integrates eye-tracking data cleaning, remapping, and hybrid field diffusion. By leveraging manifold geodesic distance as the primary constraint, the HCD algorithm ensures that saliency propagation strictly adheres to the mesh topology to achieve precise and robust saliency modeling.
In summary, the main contributions of our work are as follows:
\begin{itemize}
  \item We propose a framework utilizing VR to construct 3D mesh saliency GT. By integrating an immersive scene with enhanced methods for stereoscopic perception, we establish a data acquisition paradigm of high fidelity that aligns with natural human visual perception mechanisms.
  \item We design a VCS strategy  that mimics the receptive field of HVS. By substituting discrete intersections of single points with ray bundles weighted by probability, this approach effectively addresses spatial sparsity and discontinuity in textures of high frequency and complex topologies, enhancing the generalization and robustness of eye-tracking data acquisition for complex geometric features.
  \item We propose an HCD algorithm. By incorporating geodesic distance constraints into the eye-tracking data cleaning and remapping pipeline, we eliminate signal leakage across surfaces and topological short-circuits caused by traditional spatial smoothing, achieving precise and robust modeling of saliency GT on 3D mesh.
\end{itemize}
\section{Eye-tracking Data Acquisition}

\begin{figure*}[t]
    \centering
    \includegraphics[width=\textwidth]{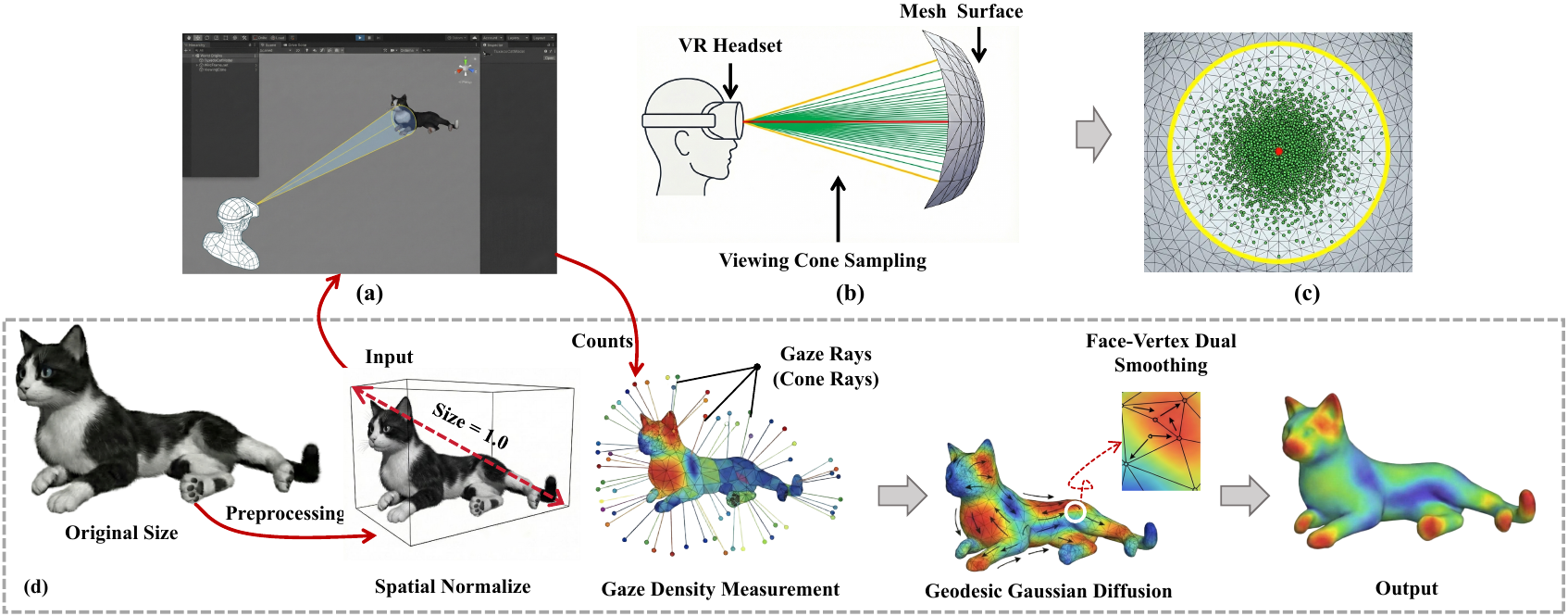}
    \caption{(a) Example of the Unity3D eye-tracking data acquisition scene. (b) Schematic cross section of the VCS strategy. (c) Example of ray distribution within the sampling field of the VCS strategy. (d) Pipeline for 3D mesh saliency GT generation.}
    \label{fig2}
\end{figure*}

In this section, we establish an immersive VR environment for eye-tracking data acquisition and utilize the VCS strategy to implicitly capture attended surface regions, thereby providing a robust data foundation for mesh saliency GT generation.

\subsection{Eye-tracking Data Acquisition Environment}

\textbf{\textit{Hardware and Software.}}
We employ an HTC VIVE PRO EYE headset ($2880 \times 1600$, $110^\circ$ FoV, $90\text{Hz}$) integrated with Unity3D and SteamVR for data acquisition.

\textbf{\textit{Scene Setup.}} 
Subjects operate within a $3\text{m} \times 3\text{m}$ 6-DoF space. We position target meshes at the origin and render a monochromatic background to eliminate visual clutter, ensuring that visual attention is driven exclusively by the mesh's saliency features. To ensure uniform surface exposure, we rotate the models at a constant angular velocity of $15^\circ/\text{s}$ for a $25s$ observation period.

\textbf{\textit{Lighting.}} 
To balance visual fidelity with low-latency requirements, we employ baked Global Illumination with static point lights and light probes. By integrating an artifact free light probe, we ensure the rotating mesh receives stable ambient lighting and enhanced geometric perceptibility\cite{lavoue2018visual}. This configuration effectively maintains high visual coherence during the interactive observation process, ensuring that dynamic light-shadow transitions remain stable and natural.

\textbf{\textit{Participants.}}
We recruited $22$ participants with normal color vision. Before the session, we provided only essential instructions regarding the equipment and experimental procedures to prevent prior bias \cite{martin2024sal3d}. Following a standard 5-point calibration, subjects performed natural visual exploration.

\subsection{View Cone Sampling Strategy}

In traditional single ray acquisition schemes, gaze ray generation typically employs a two-stage coordinate transformation mechanism. First, the eye-tracking module derives a normalized gaze direction vector within the VR headset's local coordinate system based on the Pupil Center Corneal Reflection (PCCR) algorithm \cite{guestrin2006general}. The local vector is subsequently mapped into the global Unity world space via a rigid body transformation matrix determined by the real-time 6-DoF head pose, yielding the absolute gaze direction vector. The origin of the gaze ray is determined by superimposing the VR headset's current world coordinates with the eye's position in the local VR headset coordinates. Based on this method, we propose the VCS strategy to optimize the acquisition process through the expansion of the single ray into a conical ray bundle. As shown in Figure.~\ref{fig2}(a)-(c). 

First, based on the fixation characteristics of the HVS, we model the FoV of the VR headset as a cone with the apex located at the viewpoint. The primary gaze direction emitted from the headset constitutes the axis of the cone (referred to as the central ray), and the cone's apex angle $R_f$ defines the scope of the observation region. We apply multiple rotation transformations to the central ray to derive a plurality of sampling rays, thereby forming a densely sampled ray bundle. Specifically, we define a roll angle $R_r$ and a spread angle $R_s$ in the coordinate of the Unity world. The $R_r$ represents the rotation around the central ray, which determines the specific azimuthal position of the sampling ray relative to the axis. We ensure that $R_r$ follows a uniform distribution $U\sim[0, 2\pi]$ to guaranty isotropy in all directions. The spread angle $R_s$ represents the deviation angle from the central ray, determining the eccentricity of the sampling ray on the sphere centered at the viewpoint. To simulate the characteristic attenuation of visual acuity with increasing eccentricity\cite{strasburger2011peripheral}, we employ the Box-Muller transform to generate $R_s$ according to a Gaussian distribution, as formulated in Eq.\eqref{1}:
\begin{equation}
R_s = \sigma_1 \cdot \sqrt{-2 \cdot \ln(u_1)} \cdot \sin(2\pi u_2),\quad R_s \in (0, \frac{R_f}{2}) \label{1}
\end{equation}
where $\sigma_1$ denotes the standard deviation of the Gaussian distribution and $u_1, u_2$ are two independent random variables following a uniform distribution, $u_1, u_2 \sim U(0, 1)$. Subsequently, based on $R_r$ and $R_s$, we construct the rotation matrices $M_R$ and $M_S$ to transform the central ray into the sampling ray. The direction vector $d_N$ of the generated sampling ray is given by Eq.\eqref{4}:
\begin{equation}
d_N = M_C \cdot M_R \cdot M_S \cdot d_0 \label{4}
\end{equation}
where $M_C$ represents the base transformation matrix that aligns the central ray with the world coordinate, and $d_0$ denotes the initial vector, set to $[0 \quad 0 \quad 1]^T$. We utilize Unity's physics engine for ray casting to simultaneously acquire collision information between the sampling rays and the mesh model surface. Significantly, the computational cost does not increase linearly with the number of rays. Due to the high coherence of the sampling rays within the cone, the scene can fully leverage the spatial locality optimization of the underlying physics engine. This enables to employ a large number of sampling rays to ensure acquisition quality, without compromising the real-time performance required for immersive interaction. Simultaneously, we implement a threshold filter to cull surfaces that are not facing the viewpoint or are facing the viewpoint but have a grazing angle with the sampling ray \cite{zhang2024robust}, as shown in Eq.\eqref{5}:
\begin{equation}
Inf = \begin{cases} 
1 & \text{if } n_f \cdot (-\widehat{d_N}) > 0.1 \\ 
0 & \text{otherwise} 
\end{cases}, \label{5}
\end{equation}
where $Inf$ represents the valid collision information retained after threshold filtering. $n_f$ denotes the surface normal vector of the intersected mesh face and $\widehat{d_N}$ is the normalized direction vector of the sampling ray. We set the threshold at $0.1$, which classifies the incidence angles ranging from $84.26^{\circ}$ to $90^{\circ}$ between the ray and the surface normal as grazing angles.

It is worth noting that the VCS strategy does not require additional heuristic fixation extraction or fixation filtering. Specifically, the core of VCS relies on an accumulated hit count mechanism at a fixed frequency to quantify local saliency. During continuous gaze tracking, due to the extremely short duration of human saccades, their accumulated hit frequency on the mesh surface is negligible. Consequently, this accumulation mechanism naturally acts as a low-pass filter in a physical sense. This allows the system to implicitly filter out high-frequency transient saccade noise and smoothly extract stable fixation distributions without the need for any hard truncation thresholds. The VCS strategy not only eliminates the need for additional data cleaning steps but also effectively avoids biases that may be introduced by manual filtering rules, ensuring high fidelity in the construction of saliency ground truth.

\section{Mesh Saliency GT Modeling}
We present a computational framework to transform discrete eye tracking data into continuous mesh saliency maps, as illustrated in Figure. \ref{fig2}(d). To mitigate challenges such as viewpoint randomness, sampling sparsity, and discretization artifacts, we develop the HCD algorithm. This pipeline encompasses spatial normalization, cumulative density estimation, and dual smoothing at the vertex level. By incorporating geodesic distance constraints into the data cleaning and remapping process, we eliminate signal leakage across surfaces and topological short-circuits inherent in traditional spatial smoothing, directly transforming discrete ray intersections into a continuous saliency field that faithfully aligns with the perception of the HVS and ensures robust modeling of 3D mesh saliency.
\subsection{Geometric Preprocessing and Normalization}
To ensure scale invariance and parameter consistency across models, each input mesh ($\mathcal{M} = (\mathcal{V}, \mathcal{F})$) is spatially normalized. In this context, $\mathcal{V}$ and $\mathcal{F}$ denote the vertex and face sets, respectively. Scale unification is achieved through the diagonal length $L_{diag}$ of the Axis Aligned Bounding Box (AABB), defined as $L_{diag} = \| \mathbf{p}_{max} - \mathbf{p}_{min} \|_2$, where $\mathbf{p}_{max}$ and $\mathbf{p}_{min}$ denote the maximum and minimum vertex coordinates of the AABB. We apply an isotropic scaling transformation to normalize the AABB diagonal length of all models to a unit length of $1$. This step ensures that the subsequent Gaussian kernel parameter $\sigma$ possesses relative scale invariance.

\subsection{Gaze Density Measurement}
In traditional research on 2D saliency, time decay is often introduced to simulate the recency effect of working memory. However, in 3D eye-tracking data acquisition, due to the stochastic initialization of the model loading pose, the chronological order in which users discover regions of interest is significantly confounded by random viewing angles rather than being determined purely by cognitive priority. To eliminate this systematic bias, we discard weighting methods based on time series and adopt a cumulative density invariant to time. Given that the eye tracker operates at a fixed sampling frequency, the hit count exhibits a strict linear relationship with dwell time. For any face $f_i \in \mathcal{F}$ on the mesh, its raw saliency impulse $S_{raw}(f_i)$ is defined as the cumulative hit count across all subjects during the total observation period:
\begin{figure*}[t]
    \centering
    \includegraphics[width=\textwidth]{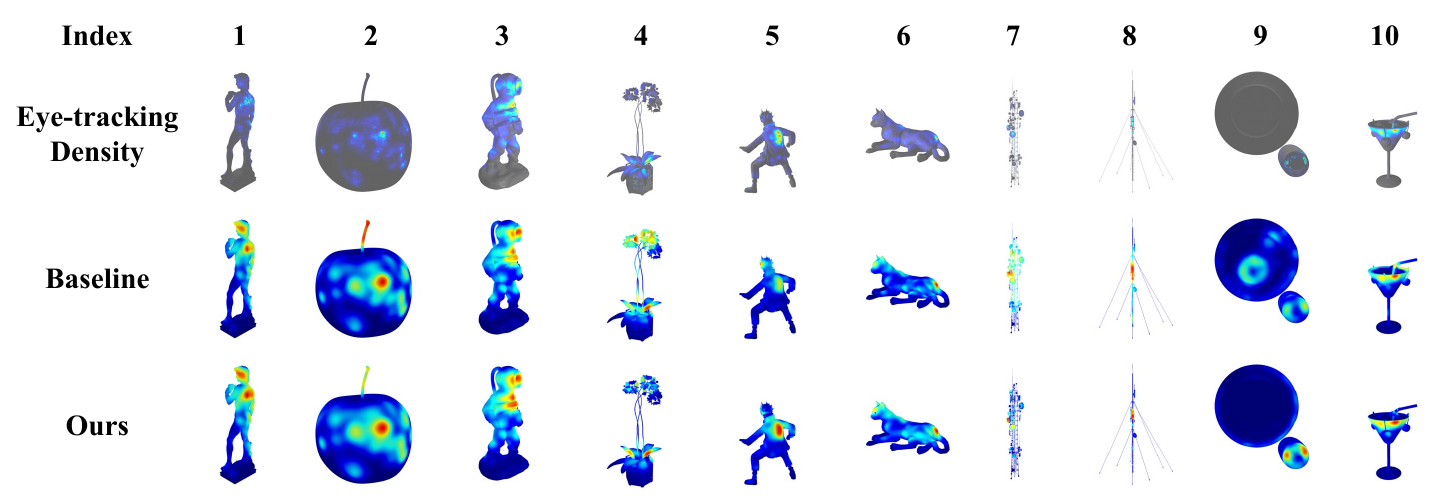}
    \caption{Qualitative comparison visualization of saliency maps on representative 3D mesh models.}
    \label{figexp}
\end{figure*}
\begin{equation}
S_{raw}(f_i) = \sum_{k=1}^{N} \mathbb{L}(R_k \cap \mathcal{M} = f_i),
\label{8}
\end{equation}
where $N$ represents the total number of sampling points recorded across all subjects, $R_k$ denotes the $k$-th gaze ray, and $\mathbb{L}(\cdot)$ is the indicator function, objectively reflecting the absolute attention captured by the region within the fixed observation period.

\subsection{Hybrid Manifold-Euclidean Constrained Diffusion}
The original $S_{raw}$ distribution exhibits extreme spatial sparsity. To recover a continuous attention field while avoiding the topological short-circuits inherent in Euclidean smoothing, we propose the HCD algorithm. This process is modeled as energy transfer across the mesh, strictly adhering to the surface manifold.
This prevents the gaze signal from violating the geometric structure of the object and causing penetration across surfaces (e.g., penetrating directly from the face to the back of the head). The diffusion process is modeled as energy transfer across the mesh\cite{crane2013geodesics, sun2009concise}. For a central face $f_c$ and an arbitrary target face $f_j$, the diffused saliency follows a Gaussian distribution based on the geodesic distance $d_{\mathcal{G}}$:
\begin{equation}
S_{diff}(f_j) = \sum_{f_c \in \mathcal{F}} S_{raw}(f_c) \cdot \exp\left(-\frac{d_{\mathcal{G}}(f_c, f_j)^2}{2\sigma_2^2}\right).
\label{9}
\end{equation}
To mitigate scale distortion caused by heterogeneous mesh densities, we implement an adaptive dynamic breadth-first search (BFS) strategy. By sampling the average topological step length, the BFS dynamically determines the search depth to approximate $d_{\mathcal{G}}$ efficiently, truncating the search space at $d_{max} = 3\sigma_2$ in accordance with the $3\sigma$ rule.
While geodesic diffusion ensures macroscopic topological correctness, the discrete nature of mesh faces inevitably introduces high-frequency aliasing. To finalize the Hybrid Manifold-Euclidean framework, we introduce a face-vertex dual smoothing strategy as the local Euclidean constraint.

\textbf{\textit{Mapping from Face to Vertex.}} Leveraging topological adjacency, we map the saliency $S_{diff}(f)$ defined on the faces to the vertex $v$:
\begin{equation}
S_{vertex}(v) = \frac{1}{|Adj(v)|} \sum_{f \in Adj(v)} S_{diff}(f),
\label{10}
\end{equation}
where $Adj(v)$ denotes the set of faces incident to $v$.

\textbf{\textit{Laplacian Smoothing.}} We apply Laplacian smoothing to vertex data as a low pass filter to suppress noise of high frequency\cite{taubin1995signal, meyer2003discrete}. The update rule for iteration $k$ is:
\begin{equation}
S^{(k)}(v) = (1-\lambda)S^{(k-1)}(v) + \lambda \sum_{u \in \mathcal{N}(v)} \frac{1}{|\mathcal{N}(v)|} S^{(k-1)}(u),
\label{11}
\end{equation}
where $\mathcal{N}(v)$ denotes the immediate neighbors of $v$. Finally, the normalized field ${S}(v)$ undergoes nonlinear Gamma correction ($\gamma=0.5$) for contrast enhancement before mapping to RGB space.

\section{Experimental Results and Discussion}
The proposed framework is evaluated on a high quality 100 textured mesh dataset (sourced from Free3D \cite{Free3D}), which spans diverse semantic categories and resolutions (1k–1,000k faces) to ensure robustness at varying levels of detail. To align with characteristics of the HVS, the cone aperture $R_f$ for VCS is set to $5^\circ$ to represent foveal vision \cite{patney2016towards}. The number of random rays for VCS is set to 100. The sampling distribution $\sigma_1 = R_f/6$ adheres to the $3\sigma$ rule for a ray concentration of $99.7\%$. For the subsequent diffusion stage, we adopt $\sigma_2 = 0.02$ to faithfully simulate the coverage of high acuity of the fovea centralis \cite{chamberlain2007eye}, accurately modeling the saliency decay relative to the gaze point. The selection of this parameter is based on the physiological hard constraints of the HVS, rather than the result of empirical tuning. In subsequent experiments, given that prevailing approaches in the mesh saliency domain generally adopt a pipeline of single ray acquisition combined with Euclidean smoothing, we adopt this paradigm as the baseline method to conduct comprehensive comparison and ablation experiments. In current state-of-the-art saliency research using VR \cite{ding2023towards,martin2024sal3d,zhang2025mesh,zhang2025textured, zhang2025unified, zhang2025elevating}, the underlying acquisition all relies on this paradigm; therefore, using it as the baseline is sufficient.

\subsection{Comparative Analysis of Qualitative Results}
As shown in Figure.~\ref{figexp}, we compare the baseline method (SRS + Euclidean smoothing) with the proposed method (VCS + HCD). While both yield comparable results for models of low complexity, the proposed approach demonstrates superior robustness as resolution and topological complexity increase. It produces cohesive saliency regions with minimal noise, aligning closely with GT density maps. Specifically, the baseline method reveals intrinsic flaws on nonconvex topologies such as \#4 (Plant) and \#9 (Plate). By relying on spatial linear distance rather than topological connectivity, it induces saliency leakage across structures that are spatially proximal but disconnected. In contrast, the proposed method enforces strict manifold constraints to ensure topological correctness. For intricate geometries like \#7 (Towers1) and \#8 (Towers2), the synergy between VCS and geometric diffusion prevents signal dispersion into voids while overcoming sampling sparsity on slender structures. This combined mechanism yields saliency maps of high quality characterized by sharp boundaries and topological integrity.
\begin{figure}[t]
    \centering
    \includegraphics[width=\linewidth]{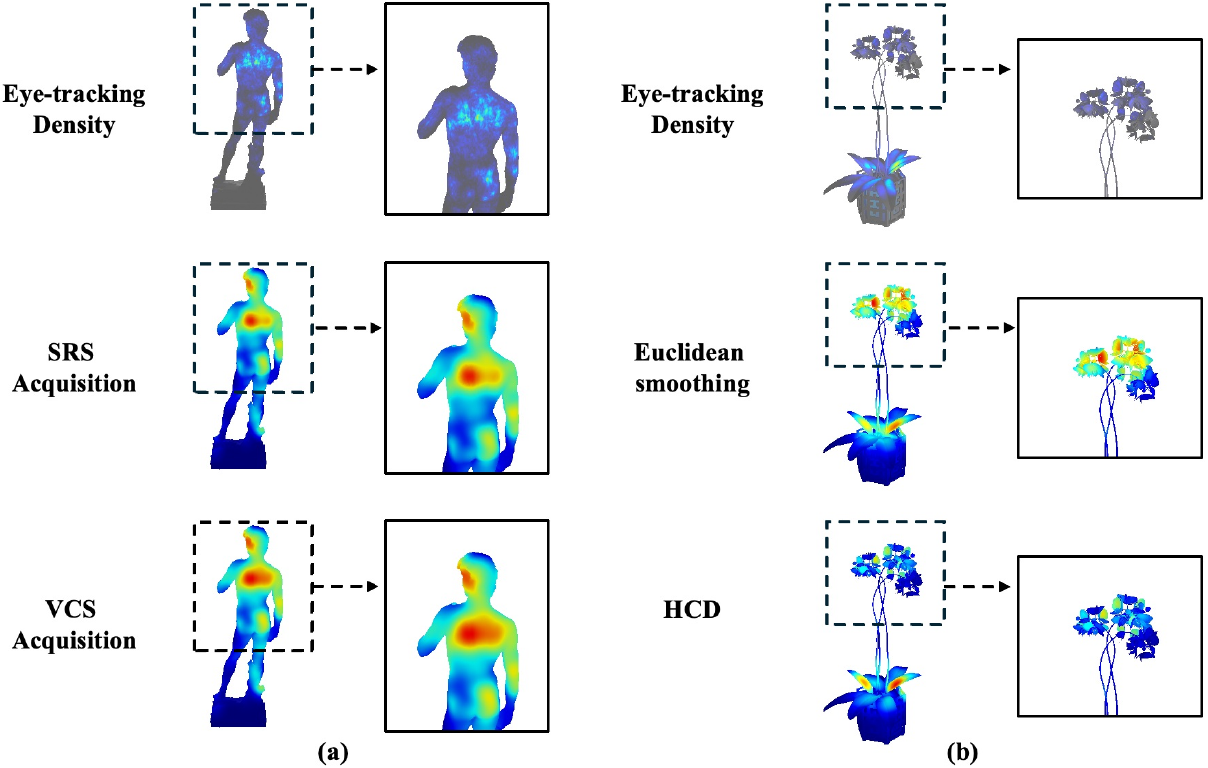}
    \caption{(a) Comparison of capture methods (Post-processing as HCD). (b) Comparison of post-processing methods (The data is sampled via VCS).}
    \label{figab}
\end{figure}
\textbf{\textit{Ablation Analysis.}} We conduct an ablation study to isolate the contributions of the VCS strategy and the proposed processing pipeline. The study is designed with two configurations: 1, Sampling Strategy: As shown in Figure.~\ref{figab}(a), we compare data acquisition using SRS versus VCS, while fixing the generation method to our proposed HCD pipeline. To ensure trajectory consistency, the single ray is defined as the central axis of the visual cone and is recorded synchronously with the VCS data. 2, Processing Pipeline: As shown in Figure.~\ref{figab}(b), we utilize VCS for data acquisition in both cases but compare the generation of saliency using the Euclidean smoothing baseline versus our proposed HCD pipeline. 

Observations indicate that data acquired via VCS effectively mitigate the sampling gaps and exhibit superior spatial continuity, preserving a saliency peak distribution that aligns highly with the eye-tracking density maps. Furthermore, the HCD pipeline restricts diffusion strictly along the manifold. This successfully prevents signal leakage across topological gaps, ensuring topological correctness of saliency propagation.

\subsection{Comparative Analysis of Quantitative Results}

In quantitative experiments, we employ Shuffled Area Under the Curve (sAUC), Correlation Coefficient (CC), and Kullback-Leibler Divergence (KL) as metrics to evaluate the correspondence between saliency maps and eye-tracking density maps. sAUC, serving as a location-based metric, effectively eliminates inherent human observation bias and evaluates the model’s spatial discriminability for real fixation points. CC focuses on assessing the linear correlation between the predicted distribution and the ground-truth density distribution in terms of overall trends. KL divergence, serving as a probabilistic distribution metric, imposes strict penalties on false negatives in saliency prediction\cite{zhang2025mesh}. Furthermore, we introduce the Internal Consistency (IC) metric to quantify the statistical stability of data obtained through different acquisition mechanisms, as defined in Eq. (\ref{eq10}):
\begin{equation}
\text{IC} = \text{CC}(\psi(E_{odd}), \psi(E_{even})),
\label{eq10}
\end{equation}
where $\psi$ denotes the saliency generation function, and $E_{odd}$ and $E_{even}$ represent the eye-tracking data sequences corresponding to odd and even frames, respectively.
\begin{table}[t]
    \centering
    \caption{Quantitative comparison of different acquisition strategies and processing pipelines}
    \label{tab1}
    \renewcommand{\arraystretch}{1} 
    \setlength{\tabcolsep}{1.5pt} 
    \begin{tabular*}{\linewidth}{@{\extracolsep{\fill}} l c l c c c @{}}
        \toprule
        \textbf{\begin{tabular}[c]{@{}l@{}}Acquisition\\Strategy\end{tabular}} & 
        \textbf{IC} ($\uparrow$) &
        \textbf{\begin{tabular}[c]{@{}l@{}}Processing\\Pipeline\end{tabular}} & 
        \textbf{sAUC} ($\uparrow$) & 
        \textbf{CC} ($\uparrow$) & 
        \textbf{KL} ($\downarrow$) \\
        \midrule
        \multirow{3}{*}{SRS} & \multirow{3}{*}{0.0557} & Direct & 0.7865 & 0.2194 & 2.8791 \\
         & & Baseline & 0.7756 & 0.1970 & 3.2092 \\
         & & \textbf{Ours} & \textbf{0.8050} & \textbf{0.2568} & \textbf{2.7753} \\
        \midrule
        \multirow{3}{*}{\textbf{VCS}} & \multirow{3}{*}{\textbf{0.8137}} & Direct & 0.7621 & 0.4571 & 1.1820 \\
         & & Baseline & 0.7709 & 0.3793 & 1.4400 \\
         & & \textbf{Ours} & \textbf{0.8288} & \textbf{0.4829} & \textbf{1.1278} \\
        \bottomrule
    \end{tabular*}
\end{table}

\begin{figure}[t]
    \centering
    \includegraphics[width=\linewidth]{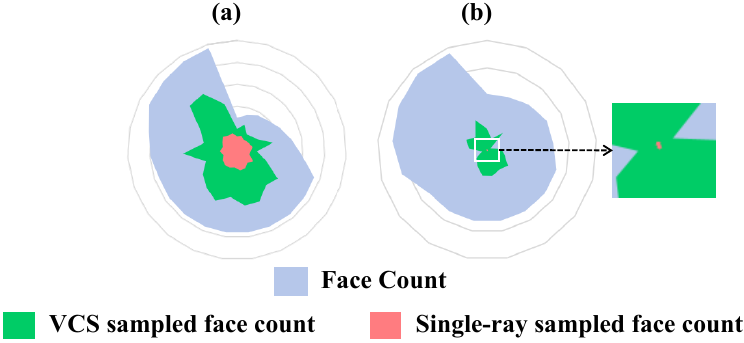}
    \caption{Statistical comparison of sampling coverage efficacy. (a) Face counts ranging from 5k to 50k. (b) Face counts ranging from 400k to 800k.}
    \label{figsam}
    \vspace{-18pt}
\end{figure}

\textbf{\textit{Ablation Analysis.}} As shown in Table \ref{tab1}, we conduct a cross evaluation comparing two acquisition strategies (SRS and VCS) across three processing methods (Direct: diffusion based on patch indices, Baseline: Gaussian smoothing based on Euclidean distance, and Ours: proposed HCD processing pipeline). Experimental results demonstrate substantial performance gains from the baseline (SRS + Euclidean Smoothing) to our proposed framework (VCS + HCD). Specifically, CC increases from 0.1970 to 0.4829 ($2.45\times$), while KL decreases from 3.2092 to 1.1278, indicating strong alignment with eye-tracking density maps. Furthermore, sAUC reaches 0.8288. This performance leap stems from the synergy between data of high reliability and geometric algorithms of high fidelity.
Our proposed pipeline demonstrates exceptional robustness. Under the sparse data conditions of SRS acquisition, our method improves CC by 30.4\% to 0.2568 and elevates sAUC to 0.8050 compared to Euclidean smoothing. These results confirm that the geodesic propagation mechanism provides strong topological completion. By enforcing manifold constraints, we effectively correct spatial errors in data of low quality to yield plausible saliency distributions. Furthermore, the transition to the VCS acquisition mechanism provides a fundamental advancement. Results indicate that the Internal Consistency (IC) for SRS is merely 0.0557, which signifies a failure to capture stable attention patterns. In contrast, switching to VCS elevates the IC to 0.8137. This consistent data provides a robust foundation for all algorithms and enhances performance across comparative methods. Building on this foundation, the “VCS + HCD” configuration achieves an optimal sAUC of 0.8288 and the minimum KL divergence of 1.1278. As corroborated by Figure. \ref{figsam}, this quality leap is attributed to the dense coverage of the VCS strategy. Table \ref{tab2} shows that these improvements become increasingly pronounced as mesh resolution increases. This synergy effectively resolves sparsity issues and bridges the gap between discrete ray casting and continuous human visual perception to ensure that the saliency maps are statistically reliable.

To simultaneously meet the demands of real-time interaction and efficient data processing, the framework is deeply optimized for computational efficiency throughout the entire pipeline. During the data acquisition stage, we leverage Unity3D’s spatial locality optimization mechanism to maintain a stable 90 Hz refresh rate even under dense VCS sampling, ensuring a zero-latency experience during immersive observation. In the post-processing stage, to address the high complexity bottleneck of geodesic calculations on high-resolution meshes, our dynamic BFS strategy based on a physical truncation threshold effectively constrains the search scale of single-point diffusion to local constant time complexity. As a result, the system achieves a high throughput of 45.25 FPS, significantly accelerating the construction of high-fidelity saliency ground truth.

\subsection{Subjective Experiment}

\begin{figure}[t]
    \centering
    \includegraphics[width=\linewidth]{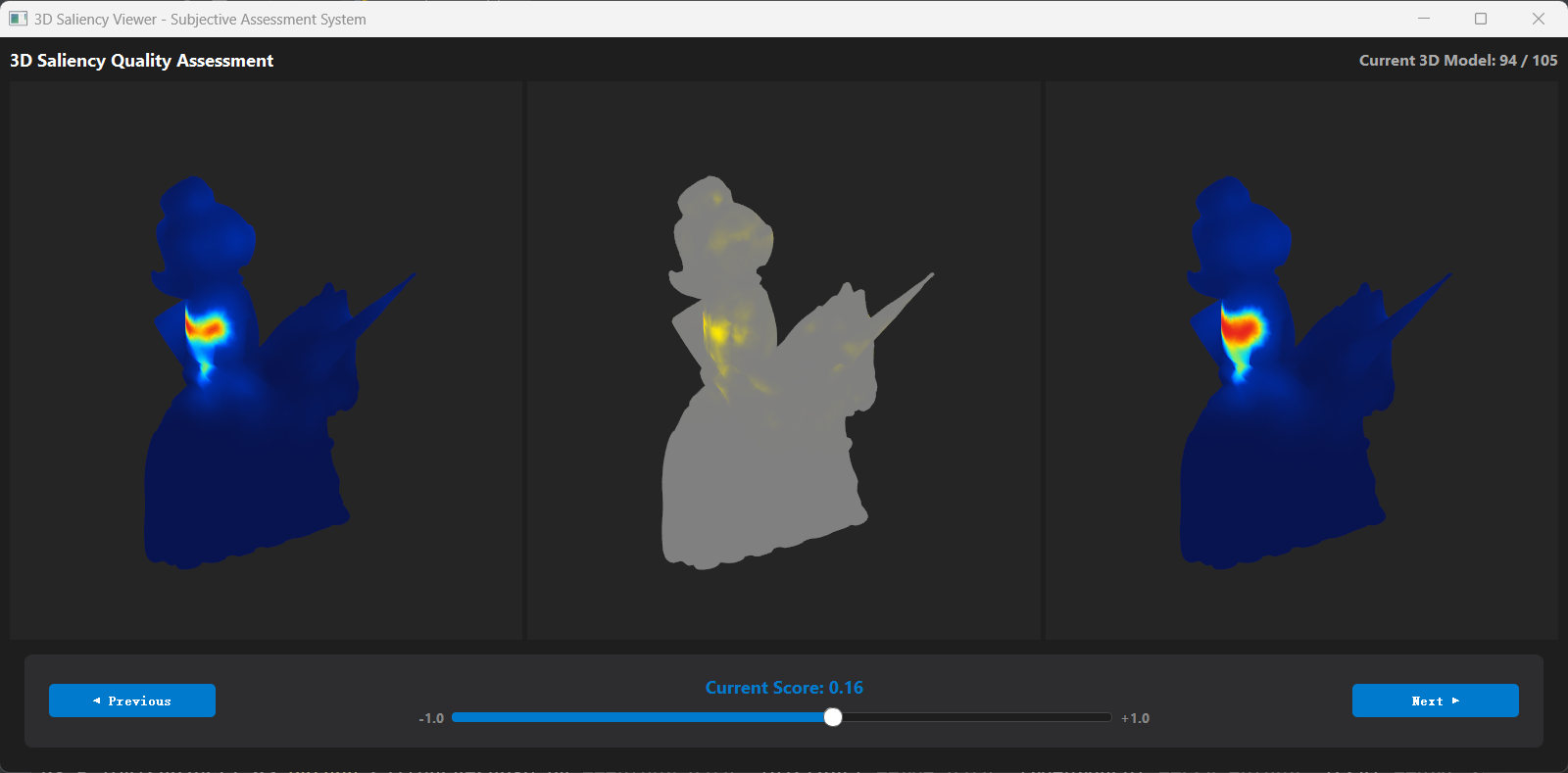}
    \caption{Subjective experimental interface. }
    \label{fig6}
    \vspace{-10pt}
\end{figure}

Although quantitative metrics provide statistical representations of data, the assessment of saliency GT construction quality should still be grounded in the HVS itself. To rigorously demonstrate the perceptual superiority of our proposed pipeline, we design a subjective 3D saliency experiment, an approach that adopts the rigorous 2D evaluation standards from ITU-R BT.500 \cite{itu_r_bt500} and ITU-T P.910 \cite{itu_t_p910}, and is built upon a modified double-stimulus continuous preference scale paradigm \cite{nehme2020visual, nehme2023textured}. Crucially, our experimental design breaks away from the static viewing constraints of traditional 2D evaluations by granting subjects full interactive freedom within a 3D space, thereby ensuring that the experiment achieves maximum ecological validity for HVS-based visual perception. As shown in Figure. \ref{fig6}. The experimental interface adopts a three‑viewport side‑by‑side layout: the central viewport displays the eye‑tracking density map as the reference; the left and right viewports present the test stimuli, where the saliency maps rendered by our proposed pipeline and the baseline method are shown in a double‑blind and randomized manner to strictly eliminate positional bias. The system forces real‑time synchronization of camera perspectives across the three viewports, allowing subjects to rotate and zoom the mesh model at will, thereby ensuring that spatial characteristics from local details to global structures can be compared. In terms of the evaluation mechanism, the experiment introduces a bipolar continuous scale ranging from [-1, +1], with the scale initially anchored at 0, indicating subjective visual equality. Subjects are required to carefully compare the match between the test stimuli on both sides and the central reference, and drag the slider toward the side that they subjectively consider to have superior saliency; the dragging amplitude reflects the intensity of the subjective preference. The design of this subjective experiment not only allows us to determine the win/loss relationship between algorithms, but also precisely quantifies the substantial improvement in saliency fidelity achieved by our proposed pipeline over the baseline model.

\begin{figure}[t]
    \centering
    \includegraphics[width=\linewidth]{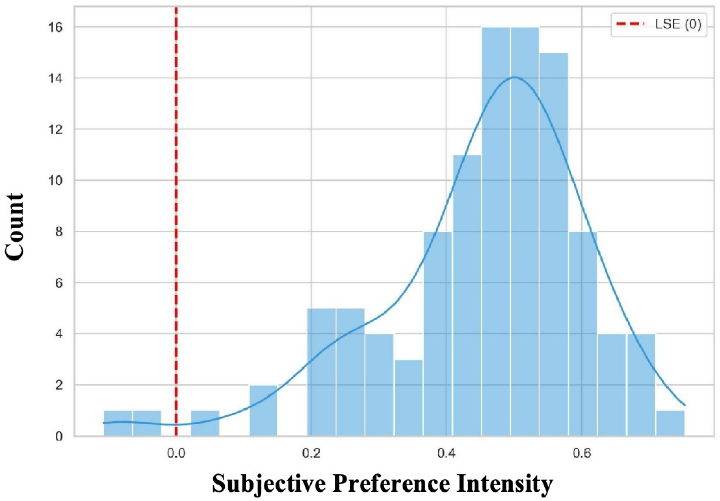}
    \caption{Statistical distribution of the subjective preference intensity. The red dashed line represents the Line of Subjective Equality (LSE = 0). }
    \label{fig7}
    \vspace{-10pt}
\end{figure}

The subjects in this subjective experiment is consistent with that in Section 2.1, and the experimental models are sourced from the dataset of \cite{zhang2025mesh}. During the post-processing stage, we performed polarity correction on the raw intensity of the subjective preference, uniformly mapping positive values to preferences for our proposed pipeline and negative values to preferences for the baseline. Figure. \ref{fig7} shows the frequency histogram and kernel density estimation (KDE) curve of the corrected subjective preference intensity. Using the line of subjective equality (LSE) as a reference, the data distribution exhibits a significant unilateral shift. Statistically, over 97\% of responses favored our proposed method. The KDE curve exhibits a unimodal structure peaking near 0.50 , with more than 70\% of preferences tightly concentrated within the [0.30, 0.65] interval. These quantitative results objectively confirm that our VCS+HCD pipeline delivers a substantial and stable improvement in human visual perception fidelity.

\begin{table}[t]
    \centering
    \caption{Statistical comparison of sampling coverage metrics across different mesh complexity levels. Improv. ($\times$) denotes the improvement factor of VCS over SRS.}
    \label{tab2}
    \renewcommand{\arraystretch}{1}
    \begin{tabular}{l c c c}
        \toprule
        \textbf{Face Count} & \textbf{VCS} & \textbf{SRS} & \textbf{Improv. ($\times$}) \\
        \midrule
        $<$100k       & 0.4650 & 0.1150 & \gainColor{13}{4.04}  \\
        100k--200k    & 0.2443 & 0.0248 & \gainColor{32}{9.84}  \\
        200k--300k    & 0.1745 & 0.0134 & \gainColor{42}{12.98} \\
        300k--600k    & 0.2741 & 0.0124 & \gainColor{71}{22.07} \\
        600k--900k    & 0.1627 & 0.0061 & \gainColor{86}{26.64} \\
        $>$900k       & 0.1150 & 0.0037 & \gainColor{100}{31.05} \\
        \bottomrule
    \end{tabular}
    \vspace{-10pt}
\end{table}
\subsection{Gain for Downstream Tasks}
\begin{table}[t]
    \centering
    \caption{Performance comparison of different pipelines on the mesh saliency prediction task. Bold represents best performance.}
    \label{tab3}
    \renewcommand{\arraystretch}{1} 
    \setlength{\tabcolsep}{1.5pt} 
    \begin{tabular*}{\linewidth}{@{\extracolsep{\fill}} l c c c c c @{}}
        \toprule
        \textbf{Method} & 
        \textbf{Pipeline} &
        \textbf{CC} ($\uparrow$) & 
        \textbf{SIM} ($\uparrow$) & 
        \textbf{KL} ($\downarrow$) & 
        \textbf{SE} ($\downarrow$) \\
        \midrule
        \multirow{4}{*}{PointNet} & SRS+Euclidean & 0.4937 & 0.5668 & 0.6591 & 0.0208 \\
        & SRS+HCD & 0.5086 & 0.5715 & 0.6320 & 0.0209 \\
        & VCS+Euclidean & 0.5356 & 0.5720 & 0.6265 & 0.0187 \\
        & VCS+HCD & \textbf{0.5722} & \textbf{0.5916} & \textbf{0.5651} & \textbf{0.0163} \\
        \midrule
        \multirow{4}{*}{PointNet++} & SRS+Euclidean & 0.5266 & 0.5816 & 0.6341 & 0.0178 \\
        & SRS+HCD & 0.5224 & 0.5667 & 0.6415 & \textbf{0.0157} \\
        & VCS+Euclidean & 0.5597 & 0.5815 & 0.6011 & 0.0163 \\
        & VCS+HCD & \textbf{0.5623} & \textbf{0.5932} & \textbf{0.5850} & 0.0159 \\
        \midrule
        \multirow{4}{*}{MeshNet} & SRS+Euclidean & 0.5287 & 0.6000 & 0.6000 & 0.0179 \\
        & SRS+HCD & 0.5570 & 0.5961 & 0.5758 & 0.0174 \\
        & VCS+Euclidean & 0.5540 & 0.6116 & 0.5629 & 0.0172 \\
        & VCS+HCD & \textbf{0.5674} & \textbf{0.6164} & \textbf{0.5421} & \textbf{0.0165} \\
        \midrule
        \multirow{4}{*}{MeshNet++} & SRS+Euclidean & 0.6855 & 0.6765 & 0.3989 & 0.0274 \\
        & SRS+HCD & 0.6857 & 0.6817 & 0.3841 & 0.0249 \\
        & VCS+Euclidean & 0.6946 & 0.6826 & 0.3889 & 0.0271 \\
        & VCS+HCD & \textbf{0.6999} & \textbf{0.6890} & \textbf{0.3827} & \textbf{0.0242} \\
        \midrule
        \multirow{4}{*}{MeshRF} & SRS+Euclidean & 0.5746 & 0.6071 & 0.5504 & 0.0184 \\
        & SRS+HCD & 0.5966 & 0.6138 & 0.5105 & 0.0165 \\
        & VCS+Euclidean & 0.5797 & 0.6090 & 0.5398 & 0.0176 \\
        & VCS+HCD & \textbf{0.5972} & \textbf{0.6140} & \textbf{0.5100} & \textbf{0.0164} \\
        \midrule
        \multirow{4}{*}{Mamba3D} & SRS+Euclidean & 0.5877 & 0.6043 & 0.5307 & 0.0170 \\
        & SRS+HCD & 0.6149 & 0.6182 & 0.5100 & 0.0161 \\
        & VCS+Euclidean & 0.5916 & 0.6055 & 0.5261 & 0.0163 \\
        & VCS+HCD & \textbf{0.6184} & \textbf{0.6175} & \textbf{0.5057} & \textbf{0.0153} \\
        \bottomrule
    \end{tabular*}
\end{table}
To verify that high-fidelity GT serves as a superior supervisory signal that fundamentally benefits downstream applications, we evaluate the practical value of our proposed pipeline by introducing it into the classic downstream task: saliency prediction for performance evaluation\cite{lezoray2025learned, wang20253d}. We utilize our proposed pipeline (VCS + HCD) and baseline method (SRS + Euclidean smoothing) alongside two cross configurations (SRS + HCD and VCS + Euclidean smoothing) to collect eye-tracking data and construct the respective saliency GTs, which is then used as supervisory signals for training and testing the prediction networks. In the experiments, we select six 3D mesh understanding and processing networks as baseline models: PointNet\cite{qi2017pointnet}, PointNet++\cite{qi2017pointnet++}, MeshNet\cite{feng2019meshnet}, MeshNet++\cite{singh2021meshnet++}, MeshRF\cite{zheng2026meshrf} and Mamba3D\cite{han2024mamba3d}. The 3D model data used in the experiments are also sourced from \cite{zhang2025mesh}, for which we re-collected eye-tracking data. Regarding the training configuration, all networks retain their default architectural settings and are uniformly trained for 50 epochs. Finally, we employ widely used evaluation metrics in saliency prediction: CC, KL, Similarity (SIM), and Saliency Error (SE, measured via MSE) to comprehensively quantify the alignment between the models' predicted saliency maps and their respective GTs constructed by different pipelines. The quantitative results are presented in Table \ref{tab3}.


As shown in Table \ref{tab3}, VCS + HCD pipeline demonstrates significant advantages over the baseline method (SRS + Euclidean smoothing), achieving state-of-the-art results across the majority of metrics. On MeshNet++, training with GT generated by VCS + HCD improves the CC metric from 0.6855 (baseline) to 0.6999, while reducing KL divergence and SE error to 0.3827 and 0.0242, respectively. On Mamba3D, this approach similarly elevates CC from 0.5877 to 0.6184, fully demonstrating the direct driving effect of high-fidelity GT on pushing the limits of model accuracy. Ablation studies highlight the independent downstream gains driven by each module. By effectively mitigating data sparsity, the VCS mechanism boosts the SIM metric to 0.5720 on PointNet. Concurrently, HCD's manifold constraints prevent topological signal leakage, raising the CC to 0.5623 on PointNet++. Ultimately, the VCS+HCD synergy yields high-fidelity supervisory signals that directly elevate the performance ceilings of downstream 3D saliency prediction models. 


\section{Conclusion}

We present a robust framework for 3D mesh saliency GT acquisition that resolves flaws in SRS and Euclidean smoothing. We introduce VCS to simulate the foveal receptive field, effectively suppressing texture aliasing. Furthermore, our HCD algorithm incorporates manifold geodesic constraints to prevent signal leakage across physical gaps. By combining qualitative, quantitative, and subjective evaluations, we demonstrate that our framework consistently outperforms baseline methods and delivers substantial performance gains for downstream tasks, establishing a paradigm of high fidelity that aligns data acquisition with natural human perception and providing a novel solution for the construction of 3D mesh saliency GT.
\bibliographystyle{ACM-Reference-Format}
\bibliography{main}










\end{document}